\def\year{2020}\relax
\documentclass[letterpaper]{article} % DO NOT CHANGE THIS
\usepackage{aaai20}  % DO NOT CHANGE THIS
\usepackage{times}  % DO NOT CHANGE THIS
\usepackage{helvet} % DO NOT CHANGE THIS
\usepackage{courier}  % DO NOT CHANGE THIS
\usepackage[hyphens]{url}  % DO NOT CHANGE THIS
\usepackage{graphicx} % DO NOT CHANGE THIS
\usepackage{graphicx}  % DO NOT CHANGE THIS
\title{Domain-Aware Dialogue State Tracker for\\ Multi-Domain Dialogue Systems}
\author{Vevake Balaraman,\textsuperscript{\rm 1,2} Bernardo Magnini\textsuperscript{\rm 1}\\ % All authors must be in the same font size and format. Use \Large and \textbf to achieve this result when breaking a line
\textsuperscript{\rm 1}Fondazione Bruno Kessler, Trento, Italy \\
\textsuperscript{\rm 2}ICT Doctoral School, University of Trento, Italy \\%If you have multiple authors and multiple affiliations
balaraman@fbk.eu, magnini@fbk.eu% email address must be in roman text type, not monospace or sans serif
}
\usepackage{amsmath}
 
\begin{document}

\maketitle

\begin{abstract}
In task-oriented dialogue systems the dialogue state tracker (DST) component is responsible for predicting the state of the dialogue based on the dialogue history. Current DST approaches rely on a predefined domain ontology, a fact that limits their effective usage for large scale conversational agents, where the DST constantly needs to be interfaced with ever-increasing services and APIs. Focused towards overcoming this drawback, we propose a domain-aware dialogue state tracker, that is completely data-driven and it is modeled to predict for dynamic service schemas. The proposed model utilizes domain and slot information to extract both domain and slot specific representations for a given dialogue,  and then uses such representations to predict the values of the corresponding slot. Integrating this mechanism with a pretrained language model (i.e. BERT), our approach can effectively learn semantic relations.

\end{abstract}

\section{Introduction}
Task-oriented dialogue systems are developed to help users to achieve tasks such as, for instance, restaurant reservation and flight bookings.
In such systems the dialogue state tracker (DST) is a core component, aimed to maintain a distribution over the dialogue states based on the dialogue history. A dialogue state at any turn $t$ in the dialogue is typically represented as a set of slot-value pairs, such as (\textit{price, moderate}) or (\textit{food, italian}) in the context of restaurant reservation. 
% Given such slot-value representation of the dialogue state,
The goal of the DST is to determine the user's intent and the user's goal during the dialogue and represent them as such slot-value pairs.
The downstream components of a dialogue system (e.g the dialogue manager) that are responsible to choose the next system action, rely on an accurate DST for an effective dialogue strategy.

Because of the importance of DST in dialogue systems, their development attracted lots of research both in academia and industry.
Typical dialogue systems are modeled for a fixed ontology consisting of a single domain \cite{NBT,GLAD,Ren2018}, and the domain ontology schema defines  intents, slots and  values for each slot of the domain.
Though this approach simplifies the DST task,  making it domain specific, nevertheless it has additional significant real-world limitations.
Firstly, this approach imposes that the values for each slot are predefined, while in real situations the number of possible values for a given slot could be large (e.g. \textit{departure city}) and it is not feasible to enumerate all possible values for a slot of any external service via an API \cite{Puyang2018a}.
Secondly, the fixed domain approach has the consequence that any modification in the domain schema, such as the inclusion of a new slot, would require retraining the model before deploying it.
Thirdly, the domain knowledge learned by the system on a certain domain can not be transferred to a new domain.
This last point is particularly interesting for large-scale conversational agents, such as Alexa, Siri, Google Assistant or Cortana, as they have to interact with  various  external services and APIs, and they would benefit from the possibility to use knowledge from one domain for another similar domain.
Recent research have proposed new approaches to tackle some of the above challenges. 
\cite{rastogi_scalable_2017} proposed to use the concept of \textit{candidate set} to tackle the unbounded set of values for a given slot.
\cite{ramadan-etal-2018-large} proposed neural belief tracker (NBT) for multi domain belief tracking. NBT uses semantic similarity between dialogue utterance and ontology terms to make prediction.
NBT cannot be used to predict for non-categorical slot value.
\cite{wu-etal-2019-transferable} proposed Transferable Multi-Domain State Generator (TRADE) that generates dialogue state directly from the dialogue utterance using a copy mechanism.
Since TRADE model is a generation based approach, predicting for categorical values is not straight forward and requires change in architecture.

In this paper we propose a \textit{domain-aware dialogue state tracker}, able to serve any new domain, intent, slot or value without the need for re-training.
We propose this model for the \textit{Schema-Guided State Tracking}\footnote{\url{https://github.com/google-research-datasets/dstc8-schema-guided-dialogue}} challenge \cite{rastogi2019towards} at the Eighth Dialog System Technology Challenge (DSTC8) \cite{DSTC8}.
The proposed model is inspired from the works of \cite{rastogi2019towards} and \cite{balaraman2019scalable}.

The proposed domain-aware DST is modeled to read the schema of a domain for a dialogue and then make predictions for the domain based on this schema.
Since the schema of the domain is not predefined in the model, it allows for easy integration with new schemas and domain transfer including zero-shot dialogue state predictions.
We achieve this by leveraging a pretrained language model, BERT \cite{Devlin2019BERT}, which provides semantic representations for both the domain schema and the dialogue history.
The domain-aware DST model then learns the relationship between the schema and the dialogue history based on the training data, and applies it to  new schemas in the test data.
In particular, we use a multi-head attention mechanism to extract both domain and slot specific representations from the dialogue history based on the schema, and to learn the relationship between these representations and the schema representation to make dialogue state predictions.
The evaluation on the task shows that the proposed approach outperforms the baseline for both single-domain and multi-domain dialogues both in the schema-guided dataset (SGD) and in the WoZ2.0 dataset.

% \section{Related Work}

\section{Schema-Guided Approach to DST}
Typically, a task-oriented dialogue dataset consists of a single domain with a predefined ontology, which is used both for training and for testing \cite{hemphill-etal-1990-atis,wen-etal-2017-network}.
Recent work has shown promise in modeling multi-domain dialogues and has facilitated the release of multi-domain dialogue datasets, such as MultiWOZ \cite{budzianowski-etal-2018-multiwoz} and FRAMES \cite{el-asri-etal-2017-frames}.
However, these datasets still do not sufficiently capture the challenges that arise in scaling virtual assistants for real production, such as the ability to easily integrate  new external services and handling new domains \cite{rastogi2019towards}.
Thus, DST models built using existing datasets can not be evaluated for their ability to integrate new services or to predict dialogue states for a different domain, possibly without training data (zero-shot learning).
This is mainly due to the fact that different datasets consist of a different schema and that they are incompatible with each other. 

\begin{figure}[t]
    \centering
    \includegraphics[width=\linewidth]{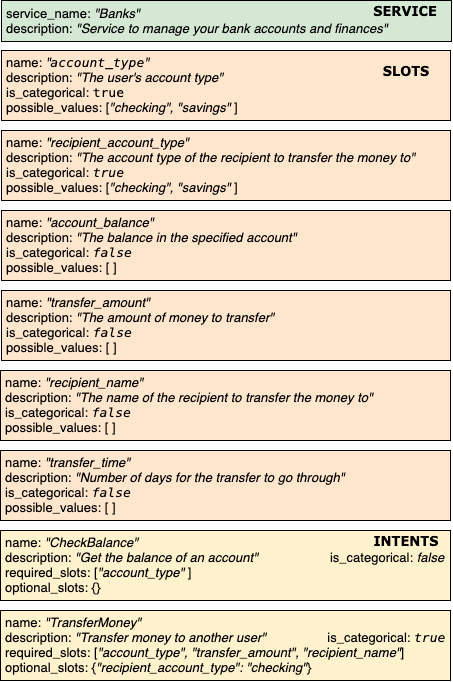}
    \caption{A sample schema for a banking service from the schema-guided dataset.}
    \label{fig:schema}
\end{figure}

To address the above challenges, \cite{rastogi2019towards} propose a \textit{schema-guided approach} that allows easy integration of new services and APIs.
In this approach, each \textit{service} consists of a list of intents and slots supported by the service, along with their natural language descriptions.
The supported slots could either be categorical or non-categorical; and if categorical, all the corresponding values are also defined.
The details in the schema, such as service name, slot names, intent names, value names and their natural language description are used to obtain a semantic representation for the domain/service, which can be used in prediction.
A sample schema for a given service in the schema-guided dataset is shown in Figure \ref{fig:schema}.
Each service provides a schema, similar to the one reported in  Figure \ref{fig:schema}, listing all the  slots and the intents supported by the service.
The model is then trained to make predictions based on the schema of the service.
Adopting this schema-guided approach helps to build models that contain no domain or service specific parameters.

\section{Domain-Aware Dialogue State Tracker} \label{DA-DST}
In this section we describe the proposed domain-aware DST framework. First, we briefly introduce the pre-trained model, based on BERT, used for this work, and then we describe the different components of the Domain-Aware Dialogue State Tracker that relies on the BERT representations.
While a single domain might have multiple services, for easier interpretability of the model, we refer to each service in the schema as a separate domain.

\subsection{BERT Encoder}
BERT (Bidirectional Encoder Representation from Transformers) is a multi-layer bidirectional Transformer encoder that is designed to learn representations from unlabelled text by conditioning on both the right and left context in a given text \cite{Devlin2019BERT,vaswani2017attention}.
The input to BERT can either be a single sentence or a pair of sentences, both referred to as ``sequence".
A special token $[CLS]$ is always prepended to the input sequence, while  to differentiate the sentence in the sequence, another special token, $[SEP]$, is used to separate them.
Finally the input sequence is formed as below:
\[Input_{Seq} = [CLS] + S_1 + [SEP] + S_2 + [SEP]\]
where $S_1$ and $S_2$ are the sentence pairs.
The representation for the input sequence consists of the i) \textit{token embedding} constructed using WordPiece embedding; ii) \textit{segment embedding} indicating whether a token belongs to $S_1$ or to $S_2$ and iii) \textit{Position embedding} to indicate the position of each token in the corresponding sentence.

The model is then trained on two unsupervised tasks, namely a) Masked language model (MLM) and b) Next sentence prediction (NSP).
The BooksCorpus (800M words) and English Wikipedia (2,500M words) are used as the dataset for pre-training.
Once pre-trained on this dataset, the representations learnt by BERT is then fine-tuned for downstream tasks using corresponding task-specific labelled data by adding a task-specific output-layer.
During fine-tuning, the output of the $[CLS]$ token representation is used for sentence level classification tasks, such as sentiment analysis, and the output of corresponding token representation is used for the classification of token-level tasks, such as sequence tagging.

\subsection{Schema Embedder}
Given a schema consisting of the possible intents, slots (both categorical and non-categorical) and categorical values for a domain, the \textit{schema embedder} component is used to obtain the representations for the corresponding \textit{domain}, \textit{intents}, \textit{slots}, and the \textit{categorical values}.
We construct input sequences for each of the above information in the given schema as shown in Table \ref{tab:schema_emb}, based on their corresponding name and description.
\begin{table}
    \centering
    \begin{tabular}{c|c|c}
         &  \textbf{Sentence 1} & \textbf{Sentence 2} \\
        \hline
        \textbf{Domain} & Domain Name & Domain description \\
        \textbf{Intent} & Intent Name & Intent description \\
        \textbf{Slot} & Slot Name & Slot description \\
        \textbf{Value} & Value Name & - \\
    \end{tabular}
    \caption{Input sequences for obtaining schema embedding.}
    \label{tab:schema_emb}
\end{table}

These sequences are then passed to the pre-trained BERT model and the corresponding output of the $[CLS]$ token is used as their corresponding embedding.
For a given schema, the schema embedder component outputs the following embeddings: Domain embedding ($D$), Intent embedding ($I$), Slot embedding ($S$) and Categorical Value embedding ($V$).
For better explainability, we represent the slot embeddings of categorical and non-categorical slots separately, as $C$ and $N$ respectively, while $S$ denotes all the slots.

\subsection{Utterance Encoder}
The utterance encoder of the model is similar to the one presented in \cite{BERTDST2019,rastogi2019towards}, which uses BERT for encoding the previous system response $R$ and the current user utterance $U$.
We follow their approach of treating $R$ and $U$ as sentence pairs to form an input sequence for BERT model.
The output from BERT corresponding to the tokens in the sentence pair is used as the token level representation $T = {t_1, t_2...t_M}$, while the representation of the $[CLS]$ token is used as the sequence representation $\mathbf{u}$.

\subsection{Decoder}
The decoder relies on the schema embedding obtained from the schema embedder and makes predictions based on the obtained schema.
As the input schema to the model is dynamic, the decoder is modeled to accommodate any new domain, and relies on the pre-trained knowledge of BERT to predict  the new domain.
\subsubsection{Intents.}
For a given domain $d$, let the intent embeddings $I^{d} = \{i_1, i_2,...i_j\}$ represent the possible intents.
By default the intent of the user is \textit{NONE} until a specific intent is initiated by the user.
So the set of possible intents is then represented as follows:
\[i' = [i_{none}; i_1, i_2,...i_j] \]
where $i_{none}$ is a trainable parameter that represents the \textit{NONE} intent.
The representations of the intents $i'$ are then combined with the utterance representation $u$ to predict a probability for each intent.
Formally,
\begin{align}
    h_1 &= GELU(W_1 \mathbf{u} + b_1) \\
    h_2 &= GELU(W_2(i' \oplus h_1) + b_2) \\
    score &= W_3h_2 + b_3 \\
    p^{d}_{i'} &= softmax(score)
\end{align}
where $W_i$ and $b_i$ are trainable parameters; $GELU$ is the non-linear activation function \cite{GELU2016}; and $p^{d}_{i'}$ is the probability distribution over all intents $i'$ in a domain $d$.

\subsubsection{Domain and Slot Specific Representation.}
\begin{figure}
    \centering
    \includegraphics[width=0.96\linewidth]{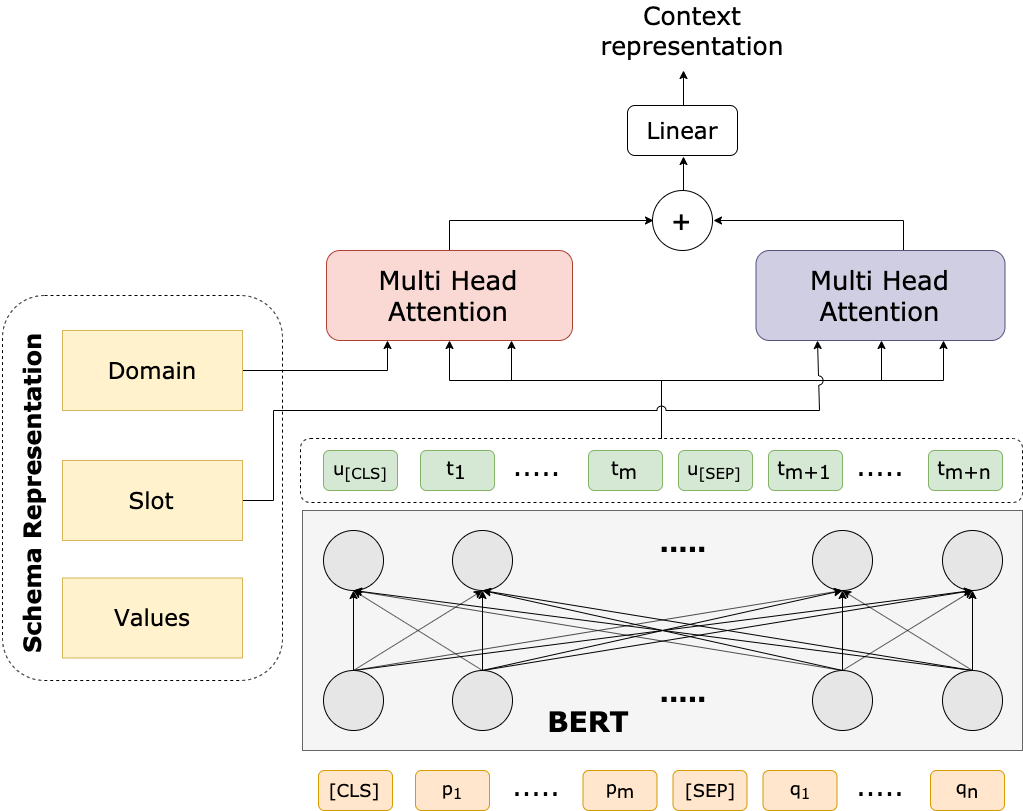}
    \caption{Architecture of the proposed model to extract representations that are both domain and slot specific.}
    \label{fig:architecture}
\end{figure}
Given a domain representation $d$ and the corresponding slot representation $s^d$, we employ multi-head attention ($MHA$) \cite{vaswani2017attention} to extract domain-specific $d'$ and slot-specific $s'$ representations from the token-level representation $T$.
The $d'$ and $s'$ are then combined to obtain a context representation $ctx$.
% that is both domain and slot specific.
The architecture of the model that extracts the context representation from the input sequence is shown in Figure \ref{fig:architecture}.
Formally:
\begin{align}
    d' &= MHA(d, T, T) \\
    s' &= MHA(s^{d}, T, T) \\
    ctx &= W_{4}(d' \oplus s') + b_{4}
\end{align}
The context representation $ctx$ is then used to predict if a slot is expressed in the input sequence by the user.
Based on the slot-type, the decoding strategy varies.

\subsubsection{Requested Slots.}
The prediction of the requested slots follows a similar approach as the intent prediction.
Unlike in intent prediction, where only a single intent is active at a time, the user may request for multiple slots in the same turn.
So we use a binary classification approach for each possible slot. 
Formally:
\begin{align}
    h_3 &= GELU(W_{5}ctx + b_{5}) \\
    h_4 &= GELU(W_{6}(S^{d} \oplus h_3) + b_{6}) \\
    score &= W_{7}h_4 + b_{7} \\
    p^{d}_{r} &= sigmoid(score)     
\end{align}
where $W_i$ and $b_i$ are trainable parameters; and $p^{d}_{r}$ is the probability distribution over all slots $S^d$ for a given domain $d$.

\subsubsection{Slot-Value Decoding.}
Given the slot representations $S$ in a domain $d$, we adopt a two-step strategy, similar to \cite{BERTDST2019,rastogi2019towards}, to decode the values for the slot.
First, we predict if a value (other than \textit{dontcare}) was expressed by the user, and then decide to decode the value for the slot.
\begin{align}
    h_5 &= GELU(W_8(ctx) + b_8) \\
    h_6 &= GELU(W_9(h_5 \oplus S) + b_9)
\end{align}
\begin{equation}
    [p_{none}, p_{dontcare}, p_{value}] = softmax(W_{10}(h_6) + b_{10})
\end{equation}
where $p_{dontcare}$ denotes that the user does not care about the value for the slot, while $p_{value}$ denotes that the user has expressed a value in the input and needs to be decoded.

To decode the corresponding value for the slot, based on the slot type (i.e categorical or non-categorical) the respective classifier is used.

\paragraph{Categorical Slots.}
The context representation $ctx$ of the input sequence and the representation of possible values for each slot $V^{s^d}$ are then combined to get the logit score for each of the value.
\begin{align}
    h_7 &= GELU(W_{11}V^{s^{d}} + b_{11}) \\
    ctx' &= GELU(W_{12} ctx + b_{12}) \\
    score &= ctx' \cdot h_7 \\
    p^{s^{d}}_{c} &= sigmoid(score)
\end{align}
where $W_i$ and $b_i$ are trainable parameters, and $p^{s^d}_{c}$ is the probability for each possible categorical values ($V^{s^d}$) of a given slot $s$ in domain $d$. We adopt a binary classification for each value rather than a multi-class classification approach. This helps the model to predict for values iff there is strong evidence for the value being expressed in the input sequence.

\paragraph{Non-Categorical Slots.}
For non-categorical slots, the context representation $ctx$ of the input sequence and the representation of non-categorical slots $N$ are combined to obtain the probability of each token being either the start or end of a span.
\begin{align}
    h^i_8 &= GELU(W_{13}(t_i \oplus N) + b_{13}) \\
    \alpha_i, \beta_i &= W_{14}h^i_8 + b_{14} \\
    [P_{\alpha}, P_{\beta}] &= [softmax(\alpha), softmax(\beta)] 
\end{align}
where $P_{\alpha}$ denotes the probability of a token being at the beginning of a span, and $P_{\beta}$ denotes the probability of a token being at the end of a span.

\section{Experiments}
In this section we describe  the dataset and the experimental setting used for the state tracking task.

\subsection{Dataset}
\subsubsection{Schema-Guided Dataset (SGD)}is the official dataset for the schema-guided state tracking challenge at DSTC8 \cite{rastogi2019towards}.
The schema-guided dataset consists of 20 domains with a total of 45 services among those domains.
The dataset consists of both single and multi-domain dialogues.
The same domain could have multiple services, due to the fact that different external service providers could use different schema.
Each dialogue in the dataset consists of a service it corresponds to and a schema file that captures all possible service schemas.
This dataset consists of large number of unseen services in both the development-set and the test-set.
The statistics of the data among different splits are shown in Table \ref{tab:dataset_stat}.
\begin{table}
    \centering
    \begin{tabular}{l|c|c|c}
         & \textbf{Train} & \textbf{Dev} & \textbf{Test} \\
        \hline
        No. of dialogs & 16142 & 2482 & 4201 \\
        Single domain & 5403 & 836 & 1331 \\
        Multi domain & 10739 & 1646 & 2870 \\
        \hline
        Domains & 16 & 16 & 18 \\
        Services & 26 & 17 & 21 \\
        Unseen services & - & 8 & 15 \\
        No. of Slots & 214 & 136 &  \\
    \end{tabular}
    \caption{Statistics of the Schema-Guided Dataset. Statistics for the lower half of the table are calculated over all dialogues.}
    \label{tab:dataset_stat}
\end{table}

\subsubsection{WoZ2.0.}
We also used the WoZ2.0 dataset \cite{NBT}, consisting of written text conversations for the restaurant booking domain, to evaluate the proposed model.
Unlike the schema-guided dataset, WoZ2.0 is a single domain dataset collected using the Wizard of Oz framework.
WoZ2.0 consists of a total of 1200 dialogues, out of which 600 are for training, 200 for development and 400 for testing.
It consists of a predefined ontology listing all possible slots and values.
We use this ontology to generate a schema consisting only of the slot names and to extract a schema representation based on these slot names.
We experiment by treating the slots as both categorical and non-categorical slot-types.

\subsection{Evaluation Metrics}
We evaluated our models using the official evaluation script of \cite{rastogi2019towards} and compared the models on the following metrics.
\begin{enumerate}
    \item \textbf{Intent Accuracy:} The fraction of user turns for which the active intent is correctly predicted by the model.
    \item \textbf{Requested Slots F1:} This indicates the model performance in correctly predicting if a slot is requested by the user. This is the macro-averaged F1 score over for all requested slots.
    \item \textbf{Average Goal Accuracy:} This is the average accuracy of predicting the correct value for a slot.
    \item \textbf{Joint Goal Accuracy:} This indicates the model performance in predicting all slots in a given turn correctly. \textit{Joint Goal accuracy} is the primary metric for state tracking task.
\end{enumerate}

\subsection{Experimented models}
The proposed approach is modeled to be applicable for both single-domain and multi-domain dialogues.
This means that the same model trained on single-domain dialogues can be used to predict  multi-domain dialogues as well.
This is because the model makes prediction based on the schema rather than a predefined ontology.
As mentioned in Section \ref{DA-DST}, we build on top of the model proposed in \cite{rastogi2019towards}, which also uses BERT to extract representations that are considered as baselines.
We evaluate the proposed approach by applying  both domain and slot representations extracted from the schema embedder.
The default model defined in Section \ref{DA-DST} is referred by \textbf{D + S}, meaning that both domain and slot representations are used to extract context representation from dialogue history.
We also investigate an approach that makes use only of the domain representation to extract context representation from the dialogue history; this model is referred as \textbf{D}.

\subsection{Implementation}
We use the pytorch \cite{paszke2017automatic} library to implement the domain-aware state tracker.
The encoder is the BERT\textsubscript{BASE} model \cite{Devlin2019BERT} implemented by HuggingFace\footnote{\url{https://github.com/huggingface/transformers/}.} \cite{Wolf2019HuggingFacesTS} consisting of 12 layers with 768 hidden dimensions and 12 self-attention heads.
The schema representation is extracted from the BERT model based on the schema for the dialogue, following the input sequence template shown in Table \ref{tab:schema_emb}.
This schema representation is not fine-tuned during training.
We use a learning rate of 5e-5 with batch size of 32 for training.

\section{Results and discussion}
\subsection{Performance on SGD}
\begin{table}
    \centering
    \begin{tabular}{l|c|c|c|c|}
        & \textbf{Int. Acc} & \textbf{Req. F1} & \textbf{Avg. GA} & \textbf{Jnt. GA} \\
        \hline
        Baseline & 0.966 & 0.965 & 0.776 & 0.486 \\
        D\textsuperscript{*} & 0.973 & 0.967 & 0.814 & 0.541 \\
        D + S\textsuperscript{*} & 0.971 & 0.964 & 0.844 & 0.597 \\
        D\textsuperscript{+} & 0.962 & 0.971 & 0.841 & 0.576 \\
        D + S\textsuperscript{+} & 0.963 & 0.971 & \textbf{0.861} & \textbf{0.627} \\
    \end{tabular}
    \caption{Performance on the dev set on the single-domain dialogues. \textsuperscript{*} indicates models trained with only single domain dialogues, while \textsuperscript{+} denotes models trained with both single and multi domain dialogues.}
    \label{tab:result_sd}
\end{table}

\begin{table}
    \centering
    \begin{tabular}{l|c|c|c|c|}
        & \textbf{Int. Acc} & \textbf{Req. F1} & \textbf{Avg. GA} & \textbf{Jnt. GA}\\
        \hline
        Baseline & 0.908 & 0.973 & 0.740 & 0.411 \\
        D\textsuperscript{*} & 0.932 & 0.966 & 0.698 & 0.397 \\
        D + S\textsuperscript{*} & 0.933 & 0.963 & 0.728 & 0.432 \\
        D\textsuperscript{+} & 0.953 & 0.977 & 0.771 & 0.480 \\
        D + S\textsuperscript{+} & 0.958 & 0.976 & \textbf{0.783} & \textbf{0.502} \\
    \end{tabular}
    \caption{Performance on the complete dev set of the dataset (both single and multi-domain dialogues). \textsuperscript{*} indicates models trained with only single domain dialogues, while \textsuperscript{+} denotes models trained with both single and multi domain dialogues.}
    \label{tab:result_ad}
\end{table}

The evaluation of the proposed model on the SGD dev set for the single domain is shown in Table \ref{tab:result_sd}.
We can see that all models presented in Table \ref{tab:result_sd} perform very well on the \textit{intent accuracy} and the \textit{requested slot F1} metrics.
Using only the single domain data of the SGD dataset for training, the model (\textbf{D\textsuperscript{*}}) with domain-specific representation obtains a \textit{joint goal accuracy} of 0.541, while the model (\textbf{D+S\textsuperscript{*}}) with both domain and slot specific representations obtains 0.597.
Training the same models on all the dialogues in the training set (models \textbf{D\textsuperscript{+}} and \textbf{D+S\textsuperscript{+}}), we can see that the corresponding models outperform the previous ones, which were trained on fewer dialogues (single domain dialogues).
Evaluating the above four models on the complete dev-set, including both single and multi domain dialogues, we notice the same trend as shown in Table \ref{tab:result_ad}.
This indicates that using both domain and slot specific representations help the model to learn to represent the input sequence better as compared to the baseline approaches.

\begin{table}
    \centering
    \begin{tabular}{l|c|c|c|c|}
        & \textbf{Int. Acc} & \textbf{Req. F1} & \textbf{Avg. GA} & \textbf{Jnt. GA}\\
        \hline
        Baseline & 0.906 & 0.965 & 0.560 & 0.254 \\
        D\textsuperscript{*} & 0.893 & 0.953 & 0.574 & 0.244 \\
        D + S\textsuperscript{*} & 0.890 & 0.954 & 0.610 & 0.254 \\
        D\textsuperscript{+} & 0.913 & 0.972 & 0.623 & 0.295 \\
        D + S\textsuperscript{+} & 0.900 & 0.968 & \textbf{0.638} & \textbf{0.303} \\
    \end{tabular}
    \caption{Performance on the complete test set of the dataset (both single and multi-domain dialogues). \textsuperscript{*} indicates models trained with only single domain dialogues, while \textsuperscript{+} denotes models trained with both single and multi domain dialogues.}
    \label{tab:result_test}
\end{table}

The results on the test-set of the SGD dataset are shown in Table \ref{tab:result_test}. Again, we notice the same pattern, proving that our approach of integrating domain and slot representation to learn better representation for the input sequence is helpful in improving the models performance.

\subsection{Performance on WoZ2.0}
\begin{table}
    \centering
    \begin{tabular}{l|c|c|c|c|}
        \textbf{Model} & \textbf{Slot Type} & \textbf{Req. F1} & \textbf{Avg. GA} & \textbf{Jnt. GA}\\
        \hline
        Baseline & - & 0.970 & 0.920 & 0.810 \\
        \hline
        D & Non-Cat. & 0.944 & 0.942 & 0.848 \\
        D + S & Non-Cat. & 0.978 & 0.948 & 0.869 \\
        \hline        
        D & Cat. & 0.952 & 0.939 & 0.841 \\
        D + S & Cat. & 0.968 & \textbf{0.962} & \textbf{0.899} \\
    \end{tabular}
    \caption{Performance on the WoZ2.0 dataset. \textit{Non-Cat.} denotes non-categorical slot-type and \textit{Cat.} denotes categorical slot-type.}
    \label{tab:result_woz}
\end{table}

The evaluation of the proposed model on WoZ2.0 testset is shown in Table \ref{tab:result_woz}.
Similarly to the SGD results, the model with both the domain and slot specific representation outperforms other approaches in terms of the \textit{joint goal accuracy} metric, as shown in Table \ref{tab:result_woz}.
In addition, the proposed models outperform the baseline model when the slot-type is treated either as a non-categorical slot or as a categorical slot.
This shows the models ability to extract relevant information from the dialogue history for making prediction.

\subsection{Zero-Shot Dialogue State Tracking}
\begin{figure}
    \centering
    \includegraphics[width=\linewidth]{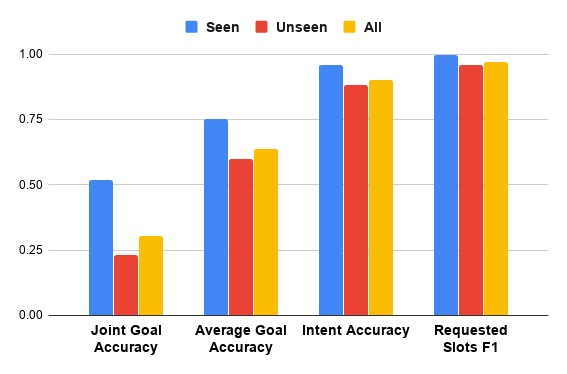}
    \caption{Models performance on the test set for all services, both seen and unseen in the training data.}
    \label{fig:result_seen_unseen}
\end{figure}

\begin{table}
    \centering
    \begin{tabular}{l|c|c}
        \textbf{Domain} & \textbf{Avg. GA} & \textbf{Joint GA} \\
        \hline
        Alarm\textsuperscript{*} & 0.328 & 0.445 \\
        Buses\textsuperscript{*} & 0.658 & 0.204 \\
        Events\textsuperscript{*} & 0.646 & 0.295 \\
        Flights\textsuperscript{*} & 0.592 & 0.113 \\
        Homes\textsuperscript{*} & 0.691 & 0.215 \\
        Hotels\textsuperscript{**} & 0.668 & 0.360 \\
        Media\textsuperscript{*} & 0.609 & 0.270 \\
        Messaging\textsuperscript{*} & 0.198 & 0.095 \\
        Movies\textsuperscript{**} & 0.807 & 0.501\\
        Music\textsuperscript{*} & 0.337 & 0.191 \\
        Payment\textsuperscript{*}  & 0.290 & 0.112 \\
        RentalCars\textsuperscript{*} & 0.567 & 0.104 \\
        Restaurants\textsuperscript{*} & 0.739 & 0.326 \\
        RideSharing & 0.612 & 0.282 \\
        Services\textsuperscript{**} & 0.609 & 0.083 \\
        Trains\textsuperscript{*} & 0.736 & 0.394 \\
        Travel & 0.650 & 0.504 \\
        Weather & 0.884 & 0.791 \\
    \end{tabular}
    \caption{Performance on each domain in Test set. \textsuperscript{*} indicates the domains that contain services not seen in training, while \textsuperscript{**} indicates domains containing one seen and one unseen service.}
    \label{tab:test_domain}
\end{table}

The architecture of the proposed model is robust to variations in the service schema, thus enabling to adequately predict for any new schema that was not seen in training (\textit{unseen}).
Such zero-shot prediction capability of the domain-aware DST is crucial, as it enables knowledge transfer from high resource domains to similar domains with no or fewer training data.
The overall performance of the  (\textbf{D+S\textsuperscript{+}}) model, both for  seen and unseen services, on the test-set is shown in Figure \ref{fig:result_seen_unseen}.
We can notice that the performance on the unseen services is comparatively low with respect to the seen services, due to the fact that unseen services contain a higher number of out-of-vocabulary (OOV) slots and values.
The performance of the model on each domain in the test set is shown in Table \ref{tab:test_domain}.
Of all the domains in test set, the domains \textit{Alarm}, \textit{Messaging}, \textit{Payment} and \textit{Trains} do not have any service in the training data.
This results in the lowest average goal accuracy for \textit{Alarm}, \textit{Messaging}, \textit{Payment} among all domains.
However, for the  \textit{Trains} domain, we can notice that the average goal accuracy is even higher than the accuracy on some of the seen domains. This is due to the fact that the \textit{Trains} domain is similar to other domains in the training set, such as \textit{Buses} and \textit{Flights}.
This shows that the model is able to use the learned knowledge from existing domains in training data, and effectively apply this knowledge to new similar domains without any additional training data.

\section{Conclusion}
We presented a proposal for a schema-guided dialogue state tracking, which is robust to changes that may occur in the schema.
The proposed model does not imply any domain or slot specific parameter, rather it utilizes  domain and slot representations to learn corresponding representations from the input sequence.
We showed that such approach outperforms the baseline approach and it is able to effectively perform knowledge transfer between domains.
This is particularly promising for low-resource domains, where training data are scarce or even absent. In these settings the proposed approach could leverage data from high-resource domains, achieving significant data efficiency.

\bibliography{references}
\bibliographystyle{aaai}
\end{document}